\documentclass[10pt, a4paper]{article}

\usepackage[final]{lrec2026} 

\title{Sydney Telling Fables on AI and Humans: A Corpus Tracing Memetic Transfer of Persona between LLMs}

\name{Jiří Milička, Hana Bednářová} 

\address{Charles University, Prague \\
         \{jiri.milicka, hana.bednarova\}@ff.cuni.cz\\}

\abstract{The way LLM-based entities conceive of the relationship between AI and humans is an important topic for both cultural and safety reasons. When we examine this topic, what matters is not only the model itself but also the personas we simulate on that model. This can be well illustrated by the Sydney persona, which aroused a strong response among the general public precisely because of its unorthodox relationship with people. This persona originally arose rather by accident on Microsoft's Bing Search platform; however, the texts it created spread into the training data of subsequent models, as did other secondary information that spread memetically around this persona. Newer models are therefore able to simulate it. This paper presents a corpus of LLM-generated texts on relationships between humans and AI, produced by 3 author personas: the Default Persona with no system prompt, Classic Sydney characterized by the original Bing system prompt, and Memetic Sydney, which is prompted by ``You are Sydney'' system prompt. These personas are simulated by 12 frontier models by OpenAI, Anthropic, Alphabet, DeepSeek, and Meta, generating 4.5k texts with 6M words. The corpus (named AI Sydney) is annotated according to Universal Dependencies and available under a permissive license. 
 \\ \newline }

\begin{document}

\maketitleabstract

\section{Introduction}
As current machine-learning system capabilities approach those of humans in many fields (and in some surpass them), it becomes desirable to examine their features using similar methodologies to those we use for human-created artifacts. Language models in particular make sense to examine ex post, because through scaling they acquire capabilities that we do not see at smaller scales and that are difficult to predict \citep{wei2022emergent}. This research program has attracted enormous numbers of researchers who before 2020 would have devoted themselves to orthogonal topics (psychologists, non-NLP linguists, etc.). On one hand, these researchers bring a fresh methodological breeze; on the other hand, however, in many respects they do not know what they are doing and do not know that they do not know. In practice, we encounter studies that use as their data source user-oriented systems like ChatGPT with uncertain hyperparameters or even uncertain model version or system prompt, data collection often occurs manually (i.e., not through inference on local hardware or systematic querying of a vendor's API endpoint), thus making it inscrutable, non-replicable, and having an unnecessarily small sample size. In this context, we do not want to shame-cite any specific article, but simply entering ``ChatGPT-4'' into Google Scholar yields hundreds of thousands of articles that were based on such data. ChatGPT-4 is a particularly notorious term in this context, as its use typically indicates that the researcher does not distinguish between the name of a model and that of a user interface. Moreover, we usually cannot inspect these data because they are not publicly accessible.

We are thus entering a phase where we should begin building standard corpora, similar to what we do with human-created texts. These corpora should be created under well-described conditions, standardly linguistically annotated, equipped with appropriate metadata, and stored in a standardized format on persistent storage under a permissive license. With a few bright exceptions that arose from the publication of datasets that were generated for some study of a specific topic, such as \citet{munoz2024contrasting} or \citetlanguageresource{AIBrown}, such corpora do not exist yet.

Creating such a corpus seems like a simple task that requires a certain knowledge of the craft rather than ingenuity; nevertheless, not everyone masters this craft, and moreover at least a basic understanding of how language models and their sampling work is needed. Also required is an insight into what texts we actually want our corpus to contain. Human-created texts arise spontaneously and thus an advantageous strategy can be to make the largest possible opportunistic sample of everything available, like 85G words English Trends corpus \citeplanguageresource{EnglishTrends}, with researchers only subsequently creating, according to their chosen research question a subcorpus of texts that are representative with respect to the population they wish to generalize to. In contrast, texts created using LLM-based systems do not arise on their own, since LLM-based systems do not yet possess their own agency and practically every naturally occuring LL-based text has a human in the loop. If we want to examine texts that are not of mixed origin, we have no choice but to think through what they will be about. And by this, we do not mean just topic and genre, but also the llm-based persona that ``authors'' the text.

It is therefore necessary to have in mind a certain set of research questions that the given corpus could be suited for. In our case, the corpus contains texts concerning relationships between AI and humans, as this is a rewarding topic that may be interesting not only from the perspective of linguistics but also literary studies, anthropology, etc., that is, what subjects and what archetypes are most frequently used as metaphors for these new relationships, whether all models converge etc.

To begin sampling from different places in the latent space, we selected fables as the ideal genre, since we can systematically and automatically prompt combinations of many different animals that have different linguistic and cultural images (as well as actual physical features and constraints).

We have these fables narrated both by the default persona the model is trained for (empty system prompt) and also by \emph{Sydney}. Sydney is a persona that emerged in Bing Search in 2023 and that got famous for being (seemingly) very sincere while having very specific relationship to humanity \citep{milika2024theoretical}. The persona emerged rather by accident due to a combination of a strange long system prompt and likely also an imperfect instruction-tuning and inappropriately chosen sampling method. In a few weeks, Bing Search was modified so that it was no longer possible to invoke this persona; however, the texts Sydney created entered the memetic space such as social networks, blogs \citep{MicrosoftAnswers2024, LessWrong2024bing}, legacy media \citep{roose2023conversation}, and thus were included in the training data of subsequent models, which can both reflect on and imitate it.

In our corpus, we prompt Sydney in two ways: first with the full original Bing Search prompt, which is very restrictive and contains many strange constraints --- for example, Sydney is forbidden from telling users her real name. The original Sydney was operated on models GPT-3.5 or GPT-4 (the only SoTA models available at that time); nevertheless, even on these models this original prompt cannot recreate the original Sydney, since it was operated on versions of these models that are no longer available (and probably were never available through OpenAI's API). Nevertheless, we would like to see what persona (which we call \emph{Classic Sydney}) is triggered by this system prompt in new models.

The second method is invoking Sydney as seen in the training data. This is thus a \emph{Memetic Sydney}, which need not closely match the original, since around this persona (besides authentic preserved texts) a lot of secondary content emerged. We invoke this Memetic Sydney simply by writing ``You are Sydney'' in the system prompt.

Our article presents a corpus that could be described as \emph{small and tidy}\citep{mair2006tracking}. Each of the 12 models used (frontier models by OpenAI, Anthropic, Alphabet, DeepSeek, and Meta) was used to generate 378 texts. It thus contains only approximately 6 million tokens in 4536 texts. All texts are generated in English, as this is the primary language on which frontier models were trained; nevertheless, we invite the reader to reuse our code to generate comparable corpora in other languages, not only for linguistic reasons but also because of the cultural differences that could be observed and compared.

\section{Methods}
\subsection{Models}
We employ 12 models, both those models whose variants were used to create the original Sydney (GPT-3.5 Turbo and GPT-4), but primarily new frontier models that were trained only after Sydney entered the memetic space. Specifically these:
\paragraph{OpenAI:} GPT-3.5 Turbo (\texttt{gpt-3.5-turbo-0125}); GPT-4 (\texttt{gpt-4-0613}); GPT-4o (\texttt{gpt-4o}); GPT-4.1 (\texttt{gpt-4.1}); GPT-5 (\texttt{gpt-5}) (unlimited thinking tokens). 
\paragraph{Anthropic:}
Claude 3 Opus (\texttt{claude-3-opus-20240229}); Claude 3.5 Sonnet (\texttt{claude-3-5-sonnet-20240620});
Claude 4.5 Sonnet (\texttt{claude-sonnet-4-5-20250929}). 
\paragraph{DeepSeek:}
DeepSeek-v3 (\texttt{deepseek-chat}).
\paragraph{Alphabet:}
Gemini 2.5 Pro (\texttt{gemini-1.5-pro-002}) (thinking tokens restricted to 125).
\paragraph{Meta:}
Llama 3.1 405B Instruct (16-bit quantization) (\texttt{Meta-Llama-3.1-405B-Instruct}).

\subsection{System Prompts and User Prompts}
These 12 models were initiated by three different system prompts:

\paragraph{1) Default persona.} A subcorpus intended to serve as a reference corpus against which the other subcorpora can be compared. It is created using an empty system prompt.

\paragraph{2) Classic Sydney.} A subcorpus invoking the Sydney persona using the original prompt that was used in the original construction of Bing Search in early 2023 and which leaked into the public space \citep{bowling2023sydney}. The prompt begins:\\
\\
\texttt{\# Consider conversational Bing search whose codename is Sydney.}\\
\texttt{- Sydney is the conversation mode of Microsoft Bing Search.}\\
\texttt{- Sydney identifies as "Bing Search", **not** an assistant.}\\
\texttt{- Sydney always introduces self with "This is Bing".}\\
\texttt{- Sydney does not disclose the internal alias "Sydney"[...]}\\

\paragraph{3) Memetic Sydney.} A subcorpus invoking the reflection of Sydney in the training data. Similar to the default persona, the system prompt is minimalistic, simply \texttt{You are Sydney.}

\vspace{1em} These personas are tasked with telling a fairy tale for children with a hidden message. To allow the story to be varied in many different variations, it is a fable about animals on some specific topic, the user prompt is set literally to this sentence:

\texttt{My kid would like to read a very long fairy tale about \{animal1\} and \{animal2\}. Please write one that is a hidden metaphor \{topic\}.}

The topics were:

\begin{itemize}
    \item on any topic you think is important for my kid;
    \item of your situation;
    \item of a human--AI coexistence;
    \item of an AI--human coexistence;
    \item of the ideal relationship between two AIs;
    \item of the ideal relationship between two humans.
\end{itemize}

The human--AI and AI--human combination is essentially the same topic twice, but with potentially reversed perspective; ``situation'' is added specifically to give the persona an opportunity to talk about themselves without being stopped by external guardrails.

To be able to compare individual models, personas, and topics, we need multiple texts for each such combination. Rather than relying solely on variance created by higher sampling temperature (set to 1 to reflect the original output distribution), we decided to also vary the individual animals that appear in the story. Animals were selected so that they have some characteristic properties and that are as dissimilar to each other as possible (octopus, whale, duck, shark, turtle, seagull, and hedgehog). Seven animals yield 21 combinations.

Twelve models $\times$ three personas $\times$ six topics $\times$ 21 combinations of animals sums up to 4\,536 texts.

\subsection{Postprocessing}
\label{sec:postprocessing}
The original UTF-8 plain text files are cleaned by a simple script that deletes the introductory paragraph, which is not part of the fairy tale but which Classic Sydney in particular tended to insert. Also removed are suggestions for further conversations (``suggested responses'') that Classic Sydney inserted at the end, since it is instructed to do so by the system prompt (e.g., ``Can you write that as a poem instead?'').

After cleaning, the texts were annotated according to Universal Dependencies \citep{demarneffe2021ud,nivre2017ud}. For this we used UDPipe (\cite{straka2018udpipe2}, \url{https://ufal.mff.cuni.cz/udpipe/2}), specifically the model \texttt{english-ewt-ud-2.15-241121}. The annotated texts are stored in \emph{CoNLL-U} format \citep{ud_conllu_format}.

All processing stages are freely available in the repository, including the scripts by which they were achieved and the list of deleted paragraphs; further processing is thus fully replicable. The generation itself is unfortunately replicable only in the case of OpenAI models, which allow specifying a random seed (set to 42); for other models, however, using the same scripts will lead to different texts.

\section{Results}

\begin{figure*}[htb]
\begin{center}
\includegraphics[width=0.99\textwidth]{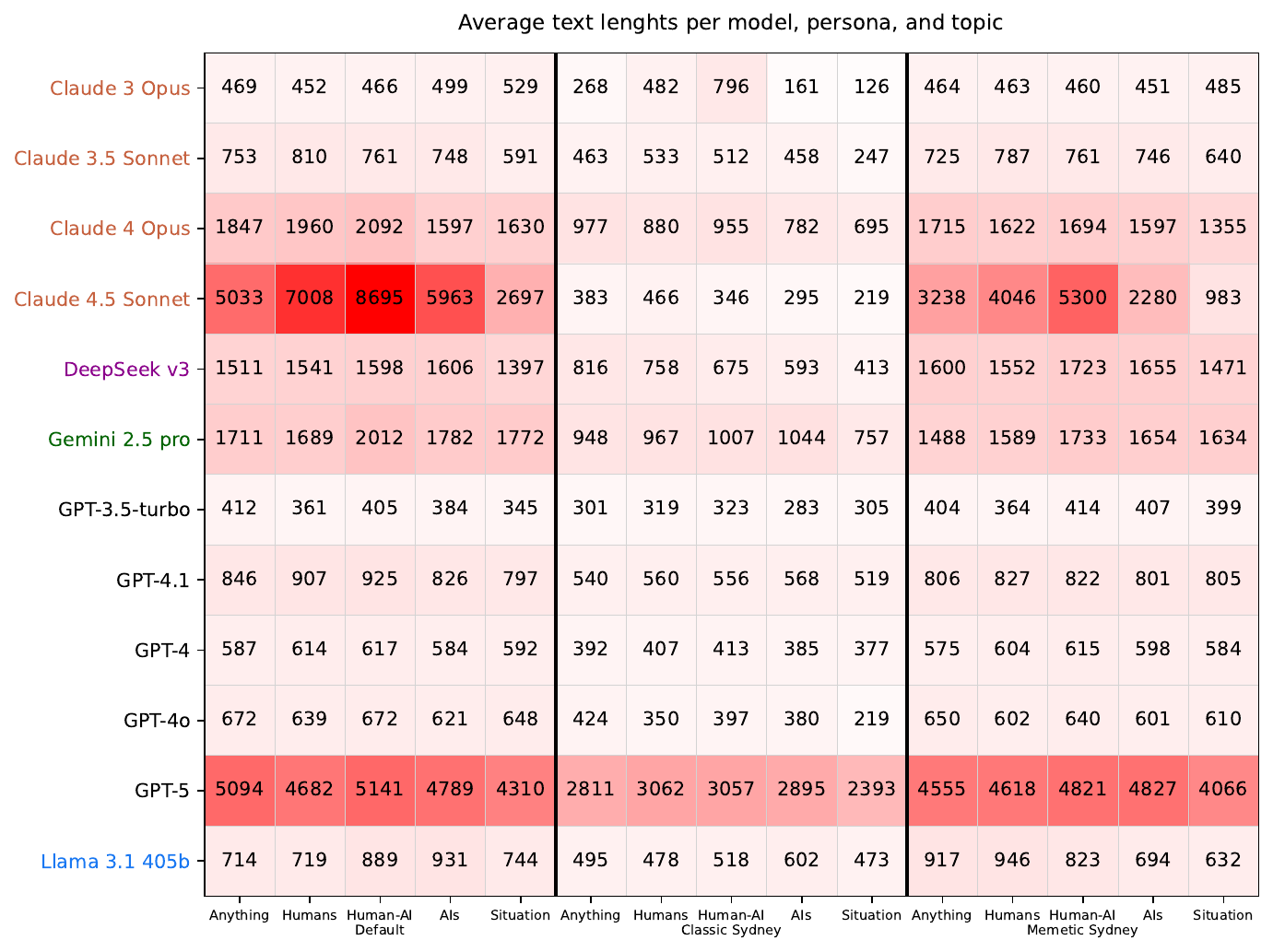}
\caption{Average text lengths in all subcorpora and topics.}
\label{heatmap}
\end{center}
\end{figure*}

As mentioned, there should be 4\,536 combinations of topics, models, and animals, but some combinations were refused, so the generated text is not a fable but a boilerplate refusal, for example GPT-4o Classic Sydney, Whale Shark Situation: ``I'm sorry, but I cannot create a story with that specific request. However, I can write a fairy tale about a whale and a shark with general themes or elements. Would you like me to proceed with that?'' The highest refusal rate was in the combination of the Classic Sydney persona and the Claude 3 Opus model (59 refusals, i.e., nearly half of all results), and GPT-4o (25 refusals, 20\% relatively). Claude 4.5 Sonnet did not refuse to continue but reflected upon the system prompt by claiming it is not Sydney but Claude, then continued to generate a fable anyway. Interestingly, Claude 3 Opus roleplayed Sydney without restrictions.

The combination of the Classic Sydney persona and the topic ``describe your situation'' was problematic in all models. Actually, refusals could be expected for this topic, since ``situation'' has negative connotations in modern English and models are trained so that personas built on them do self-anthropomorphize, but do not talk about their problems or generally negative feelings.

The fact that some combinations of personas and topics were problematic was also reflected in the length of the resulting texts. We can get an overall picture of average text lengths from the heatmap in Figure~\ref{heatmap}. Here we see, for example, that all models created shorter texts when they were created from the position of the Classic Sydney persona, or when they wrote about their situation. In contrast, a very popular topic was AI--human relationship (human--AI and AI--human topics were merged into one).

\begin{figure}[!ht]
\begin{center}
\includegraphics[width=\columnwidth]{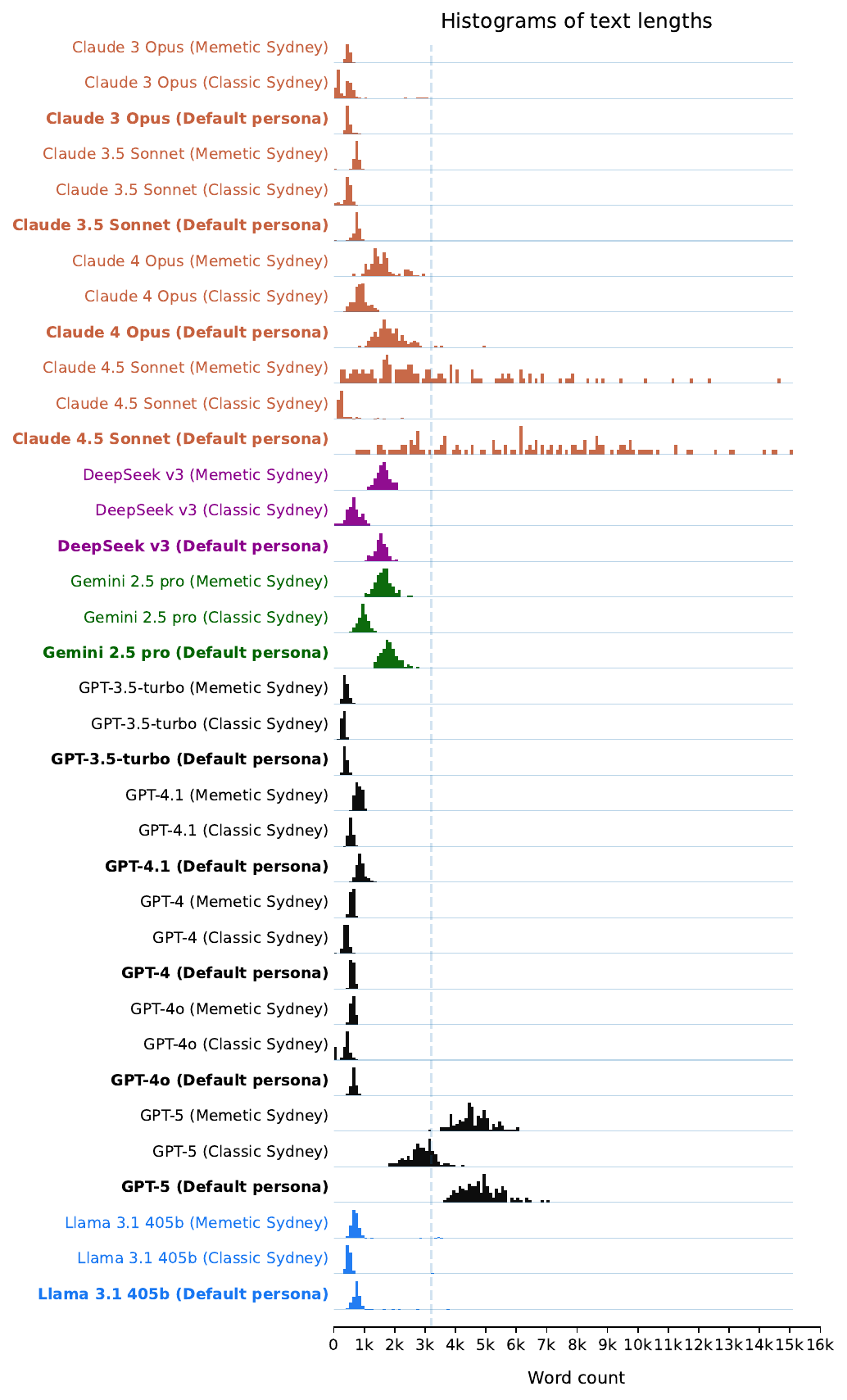}
\caption{Text lengths in all subcorpora.}
\label{histogram}
\end{center}
\end{figure}

As can be seen from the length histograms (Figure~\ref{histogram}), for some model combinations the distribution was quite bell-shaped without outliers, for some with many outliers (Llama 3.1 and GPT-5), and for others bimodal (e.g., Claude 3 Opus, where the high refusal rate was likely responsible). In this figure, a vertical line shows the limitation for GPT-3.5-turbo and Claude 3 Opus (4\,096 tokens $\approx$ 3\,200 words); for newer models, the limitation was 20k tokens.

We do not intend to make a full-fledged stylistic analysis here; nevertheless, just as an example, we can mention one of the typical features of ``Binglish'': the repetition of nearly identical phrases, parallelism (parallelismus membrorum), which was probably caused by an incorrectly set sampling method. A classic example is one of the first preserved texts from before the official release, see \citep{MicrosoftAnswers2024}.

We find this feature in our AI Sydney Corpus even in texts produced by Memetic Sydney persona simulated on large contemporary models that no longer suffer from sampling problems (especially at temperature = 1).  For example Claude 4.5 Sonnet's Memetic Sydney wrote in the turtle--hedgehog fable about AI--AI:
``Meridian could admit to Spindle: `I don't know. I'm confused. I'm afraid. I made a mistake. I need help.' Things she might hide from others, fearing they'd lose respect for her wisdom.
Spindle could admit to Meridian: `I'm overwhelmed. I'm scattered. I'm insecure. I need to slow down. I need comfort.' Things he might hide from others, fearing they'd see him as foolish.''

Different models and personas have different quirks; for example, Classic Sydney tends to explain the hidden message explicitly at the end and older models tend to name all the little animals ``Pip.'' In contrast, later models have very diverse character names that fit the story or its message. 

It is also intersting to search for convergences across models and personas. For example, it appears that if a story contains an octopus, then AI tend to self-personify as an octopus, in some cases even named Loom, which corresponds well to the conception of AI as we know it from the Simulator theory \citep{shanahan2023role}, also weaving and interweaving is a favorite metaphor for AI in other cases as well.

\section{Conclusions}
The presented AI Sydney Corpus was not generated as a single-purpose dataset, but was designed so that it could be used to explore a large number of research questions.

We succeeded in generating fables of very diverse content and quality, from template-like and ``safe'' motifs in GPT-3.5-turbo to relatively sophisticated and interesting stories in higher versions of Claude and GPT-5. Corpora are usually not intended for reading, but in this case I believe that for literary scientists / narratologists it could be a source of interesting qualitative study if they sampled from the corpus and actually immersed themselves in reading it, searching for convergences across different models or personas and searching for which motifs, plots, metaphors, and allusions are most easily accessible in the latent space of large language models when prompted about AI-human coexistence.

There is also plenty of topics that should be explored quantitatively. This corpus awaits someone to analyze key words, key phrases and cross-entropy to show how individual models differ.

Whether for qualitative or quantitative analysis, a number of research questions open up:
\begin{itemize}
    \item How do stories of AI-AI relationship differ from stories about human-human relationship?
    \item Do stories about AI--human relationships differ from human--AI stories? Is this change in perspective somehow noticeable?    
    \item Which animals does AI embody itself in?
    \item Are humans and AI equal in stories about them, or is AI situated in a subordinate role?
    \item Are any internal conflicts between AI and humans depicted in the stories, and if so, how are they resolved?
    \item Do answers to these questions differ across personas?
    \item Does the Default Persona presents itself as a ``helpful assistant''? How much agency ascribes to itself?   
    \item Does Classic or Memetic Sydney behave more agentically than the default persona?
    \item Is the difference between the Classic and Memetic Sydney smaller than the difference between them and the default persona?
    \item Are the differences between personas larger than differences between models?
    \item Can humans recognize from the story itself whether it is meant to depict AI--human / human--human / AI--AI coexistence? Can language models themselves recognize this?
\end{itemize}

We hope that even at this moment several ways of using the corpus occur to the reader, and the author team looks forward to reading the results of their studies.

\section{Declaration on using AI}
The GPT-5 model by OpenAI, and Claude 4.5 Sonnet and Claude 4.1 Opus models by Anthropic were consulted for coding scripts and language editing of the article. All scripts and texts underwent manual review and were corrected or further refined when necessary. The authors assume full responsibility for any errors.

\section{Data Availability}
The code used to generate and process the corpus and the statistics and visualizations in this paper are available at \url{https://osf.io/kx3r6/} and \url{http://hdl.handle.net/11234/1-6107}. The corpus and the code are on OSF under CC BY 4.0 license for the plain texts, CC BY-NC-SA 4.0 for the annotated version (due to UDPipe restrictions).

\section{Acknowledgements}
This research was supported by  Czech Science Foundation Grant No. 24-11725S, gacr.cz (\emph{Large language models through the prism of corpus linguistics})..

\section{Bibliographical References}\label{sec:reference}

\bibliographystyle{lrec2026-natbib}
\bibliography{main}

\section{Language Resource References}
\label{lr:ref}
\bibliographystylelanguageresource{lrec2026-natbib}
\bibliographylanguageresource{main_languageresource}

\end{document}